\documentclass[11pt,a4paper,twoside]{llncs}
\usepackage{amsfonts, indentfirst}
\usepackage{ifthen,amssymb, graphicx}
\usepackage[all]{xypic}
\usepackage{young}
\usepackage{graphicx}
\usepackage{tikz}
\usetikzlibrary{automata}

\newcommand{\cqd}{\hfill$\Box$}

\begin{document}

\title{ Modeling languages from graph networks}

 \author{Cristina Mart{\'\i}nez, Alberto Besana}
\institute{%Max Planck Institute for Mathematics, Vivatgasse 7, Bonn, 53111, Universitat,
%Departament de Matematiques, Edifici C, Facultad de Ciencies, 
%IIIA-CSIC, Campus de la UAB, 
%E-08193 Bellaterra, Catalonia, Spain}
%School of Mathematical Sciences
%UCD Dublin
%Claude Shannon Institute,
UCD CASL, University College Dublin,
8 Belfield Office Park,
Beaver Row, Clonskeagh, Dublin 4
 \email{cristina.martinezramirez@ucd.ie}
\and Fieldaware, 88 Lower Leeson St, Dublin, Ireland
\email{alberto.besana@fieldaware.com}
 }

 \maketitle

 \begin{abstract} We  model and compute  the probability distribution of the letters in random generated words in a language by using %the theory of set partitions, Young tableaux and correspondence graphs.%
 combinatorial methods and graph theoretical representation methods. 
 This has been of interest for several application areas such as network systems, bioinformatics, internet search, %search in genome data, 
 data mining and computacional linguistics. 
  \end{abstract}
 \section{Introduction}
 
 We will study languages as subsets of a monoid $A^{*}$ for a given alphabet A by means of the sequences of letters which give rise to the different words of a language. Furthermore we are interested in studying the equidistribution of letters in a given word by means of its probability distribution. 
Languages describe networks where the vertices constitute the alphabet and the transformations describe the rules. Every finite path in the graph describes a possible word in the language (\cite{OK}).
In general, languages modeled on an alphabet of 4 letters represent phenomena which occur such as multiple agreements, crossed agreements and replications. %Languages modeled on an alphabet of 2 letters.

%We are particularly interested in
 Computational linguistics has been applied in areas such as natural language interaction and computational complexity of natural language largely modeled on automata theory and universal networking language (UNL). The latter is a formal language which represents semantic data extracted from natural language texts. The universal words represent concepts and are annotated with attributes representing context information and the underlying relations between words in an existing language are represented by semantic links. The words together with the semantic links between concepts, constitute a semantic network.  From a mathematical point of view, a network consists of topology, graphs, matrices and functionality which concerns navigation. Methods of network coding consists of partitioning the network graph into subgraphs through which the same information flows, (\cite{BRJ}). The fundamental quantity describing a class of networks is the joint probability distribution, a formula encoding the probability that a randomly chosen node in the network has a given degree $k$ and is a member of a certain subgraph known as $c-$clique, a subset of the vertex set in which every two vertices in the subset are connected by an edge. The opposite of a clique is an independent set.

 Here we are interested in formal languages generated using simple grammars, %rather than natural languages, 
for example languages modeled on automata theory with application of context-free grammars. The language generated by a context free grammar is the set of terminal symbols that can be derived starting from the start symbol. The first efficient algorithm to perform a uniform random generation of words of length $n$ of a context-free grammar involved $n^{2}\,(ln\,n)^{2}$ operations on average via fast Fourier transform \cite{HC}. In the following, we present some probabilistic models which describe the probability distribution of random generated words in a language using knowledge of set partitions.
%In computer sciences languages are described as synthactic monoids. 
 In section \ref{section2} several examples of formal languages induced from free-grammars are presented and related to combinatorial objects. In section \ref{section3.1}-\ref{subsection3.4} we study a correspondence between set partitions and graphs related to words.
 
Throughout this text, a monoid is a (multiplicatively written) commutative semi-group $A$ with a neutral element 1. A morphism $f:A_{1}\rightarrow A_{2}$ of monoids is a multiplicative map with $f(1)=1$. The semiring $\mathbb{N}[A]$ is defined as the (additive) semi-group of all finite formal sums $\sum\,a_{i}$, where the $a_{i}$ are elements in $A$, possibly occurring multiple times in the sum. 

 \section{Examples of languages over an alphabet}\label{section2}% on two letters $A=\{a,b\}$}

 Let $A$ be an alphabet $\{a_{1}, a_{2}, a_{3},\ldots\}$, where $a_{i}$ is given by the weight $i$, that is a finite set containing the inverse of each element. A word (over $A$) is an element of the free monoid $A^{*}$ on $A$, that is, a finite sequence 
 \begin{eqnarray}\label{operationproduct}
(a_{1},a_{2},\ldots, a_{n})  \\
a_{1}\,a_{2}\ldots a_{n}
\end{eqnarray}

\noindent of elements of $A$. We shall consider words in a fixed alphabet $A=\{a_{1}^{\pm}, a_{2}^{\pm},\ldots\}$ of letters $a_{1}, a_{2}\,\ldots$ and their inverses $a_{1}^{-1},a_{2}^{-1},\ldots$ %where $\tau(a_{i})=a_{i}^{-1}$ and $\tau(a_{i}^{-1})=a_{i}$ is a point free involution 
and $A^{*}=\{a_{1},a_{2},\ldots\}$. A cyclic word (over $A$) is the set of cyclic permutations $[w]$ of a word $w$. A word is said to be reduced if it has no factor of the form $aa^{-1}$ or $a^{-1}a$, where $a\in A^{*}$.

 Let $A^{+}$ be the set of all these sequences over $A$, and $A^{*}$ the free {\bf monoid} $A^{*}=A\cup \{1\}$. $A^{+}$  with the concatenation product is a free semigroup. Any subset of $A^{*}$ defines a language. 

 Observe that an alphabet $A$ admits an interpretation as an unlabeled class $\mathcal{A}$ of combinatorial objects with counting sequence $\{a_{n}\}$, where the sequence operation $SEQ(\mathcal{A})$ is the class $\epsilon+\mathcal{A}+\mathcal{A}\times \mathcal{A}+\mathcal{A}\times \mathcal{A}\times \mathcal{A}+\cdots$. As the concatenation operation, it is used to define formal languages (sets of strings). In combinatorial or probabilistic terms we can look at the sequences $SEQ_{k}(A)$ of $k-$elements from a set $A$, and the generating function for these is $A(z)^{k}$. If $A(z)$ is the ordinary generating function that enumerates $\mathcal{A}$, then $\frac{1}{1-A(z)}$ is the ordinary generating function enumerating $SEQ(\mathcal{A})$. $SEQ_{k}(\mathcal{A})$ is the class of $k-$sequences of elements of $\mathcal{A}$ and $A(z)^{k}$ is the corresponding EGF enumerating $SEQ_{k}(\mathcal{A})$.

Let $K, L$ be two languages. We define the following operations on languages:
 \begin{enumerate}
 \item{Union}: $(K,L)\rightarrow K\cup L =\{u\in K \ \  or\,  u\in L\}$
 \item{Intersection}: $(K,L)\rightarrow K\cap L=\{u\in K\  {\rm and}\ v\in L\}$
 \item{Complementation}: $L\rightarrow A^{*}\ L=\{u\in A^{*}|\, u\neq L\}$
\item{Quotient}: $K\backslash L=\{u \in A^{*}|\, Ku \in L\}.$
 \item Product: $(K,L)\rightarrow K\,L=\{u\,v|\, u\in K, {\rm{and}}\, v\in L\}$
 \item Star $L\rightarrow L^{*}=\{u_{1},\ldots, u_{n}\in L, n\in \mathbb{N}_{0}\},$ that is, the submonoid of $A^{*}$ generated by $L$.
 
 \end{enumerate}
A context free grammar (CFG) is a way of describing languages by recursive rules called productions. Context free grammars were originally conceived by N. Chomsky as a way to describe natural languages (\cite{FN}). A CFG consists of a set of variables, a set of terminal symbols, and a start variable $S$, as well as the productions. Each production consists of a head variable and a body consisting of a string of zero or more variables and/or terminals. Grammars are called context free because all rules contain only one symbol on the left hand side that is the "context", in which a symbol on the left hand side of a rule occurs. For example, consider the following languages over an  alphabet on two letters $A=\{a,b\}$:% generated by the rules: 
 
 $$\emptyset$$ %\{1\}, A, A^{+}, A^{*}$
 %\item 
 $$\{1,a,b,aba, a^{8}, aabbbab\}$$
% \item 
$$\{a^{n}b^{p}|\, n,p \in \mathbb{N}\}$$
% \item 
$\{a^{n}b^{n}|\, n\in \mathbb{N}\}\backslash \{\}$, is the language consisting of a block of $a's$ followed by a block of $b's$ of equal length, except for the empty string.% and if we add the empty string $\epsilon$, we generate the language $\{a^{n}b^{n}\}$.

% \end{enumerate}
Other languages of interest in modeling natural languages are defined on alphabets of 3 letters and 4 letters:
$$L_{1}=\{a^{n}b^{n}c^{n}|\, n\geq 1\},$$
$$L_{2}=\{a^{n}b^{m}c^{n}d^{m}| \, m,n\geq 1\}.$$
These elements represent phenomena which occur in natural languages such as multiple agreements, crossed agreements and replications.
Moreover, one can study these variables by its numerical value and then we are interested in the relations between the variables for applications in cryptography. Take as an alphabet $\mathcal{P}=\{P_{1}\ldots, P_{N}\}$ the $\mathbb{F}_{q}-$rational points lying on an elliptic curve defined over $\mathbb{F}_{q}.$ Then Eva and Bob agree on a key $P_{1}\in \mathcal{C}$ and then Eva sends $P_{1}+P_{2}=P_{3}$ to Bob, so Bob knows $P_{2}=P_{3}-P_{1}$.   Or as linguistic values, variables whose values are not numbers but words or sentences in a natural or artificial language. Applications exist in Natural Language Interaction (NLI), speech recognition and universal networking language (UNL), where information is represented sentence by sentence as a hypergraph composed of a set of directed binary labeled links (referred to as relations) between nodes or hypernodes.      
 The words together with the semantic links between concepts, constitute a semantic network.
  Given two alphabets $Y=\{y_{1},y_{2},\ldots, \}$, $U=\{u_{1}, u_{2}, u_{3},\ldots\}$ one can consider the polynomial ring $\mathbb{Q}[Y]$ and $\mathbb{Q}[U]$ generated by the $u_{i}$ and the $y_{i}$, or the polynomial ring $\mathbb{Q}[U,Y]$ and then study polynomial relations (identities) between the variables $uv=vu, u=u^{2}, uv=uvu$. Then for example $uv*u$ is a set of words containing the words beginning in $u$, ending in $u$ and containing the letter $v$,  $n$ times. This is often used as a way of representing spaces of data sets and then the variables represent for example the attributes or categories attached to certain classes of data sets. The relation between the variables define classifiers covering different types of areas of the complexity space of data sets.

\begin{definition} A pattern of length $n$ admitted by a numerical semigroup $S$ is a polynomial $p(x_{1},\ldots, x_{n})$ with non-zero integer coefficients, such that, for every ordered sequence of $n$ elements $s_{1}\geq \ldots, \geq s_{n}$ from $S$, we have $p(s_{1},s_{2},\ldots, s_{n})\in S$.
\end{definition}

One can consider for example the Euclidean space $S^{n}$ of $n\times n$ symmetric matrices with inner product $\langle x, y\rangle=tr\,(xy)$. %When does a triple $(\lambda, \mu, \eta)$ of partitions
We call a triple $(\lambda, \mu, \eta)$ of eigenvalues admissible for the Horn's problem, if  they correspond to eigenvalues of real symmetric matrices $A, B, C$ such that the sum of any two of them is equal to the third.

Then the polytope of matrices admissible for  Horn's problem %admitting a triple $(\lambda, \mu, \eta)$ of partitions as eigenvalues of $n\times n$ is a numerical semigroup 
admits the homogeneus linear pattern $S_{0}+S_{1}-S_{2}$.
%Introduce definition of pattern

 \begin{definition} $Rat(A^{*})$ is the smallest set of languages over $A$ which has the emptyset and the languages $\{a\}$, with $a\in A$, and is closed under the operations of union, product and star.
 \end{definition}
 
 Computational linguistics has been applied for languages modeled on automata theory. An automata $A$ is a free and finitely presented lattice.
          
 \begin{definition} We say that a language $L\subseteq A^{*}$ is recognizable if $L=L(\mathcal{A})$ for some finite automaton. A language $L\subseteq A^{*}$ is recognizable if and only if it is rational.%due to a theorem of Kleene
 \end{definition}
 
 For example, consider the finite automata:
 
    
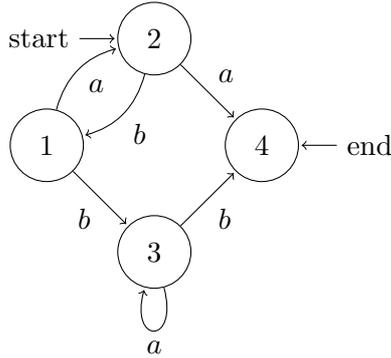
\begin{tikzpicture}[shorten >=1pt,node distance=2cm,auto]

\node[state] (1) {$1$};
\node[state,initial] (2) [above right of=1] {$2$};
\node[state] (3) [below right of=1] {$3$};
\node[state,initial,initial where=right,initial text=end] (4) [below
right of=2] {$4$};

\path[->] (1) edge node [swap] {$b$} (3)
              edge [bend left] node [swap] {$a$} (2)
          (2) edge node {$a$} (4)
              edge [bend left] node {$b$}  (1)
          (3) edge node [swap] {$b$} (4)
              edge [loop below] node {$a$} ();

\end{tikzpicture}

Observe that $L(\mathcal{A})=((ba)^{*}b)^{*}+((ba)^{*}a)^{*}$.

Words recognized by $A$ are: $a$, $baa$, $b^{3}$, $bba^{3}b$. %$L(\mathcal{A})=((ba)^{*}b)^{*}+((ba)^{*}a)^{*}$.

 \section{From $\lambda$ partitions to generating words} \label{section3}

 \subsection{Correspondence between words and partitions}\label{section3.1}
 For $d$ a positive integer,  $\alpha=(\alpha_{1},\ldots, \alpha_{m})$ is a partition of $d$ into $m$ parts if the $\alpha_{i}$  are positive and non-decreasing. We set $l(\alpha)=m$ for the length of $\alpha$, that is the number of cycles in $\alpha$, and $l_{i}$ for the length of $\alpha_{i}$. We can label these partitions by words composed of letters where the first occurrence of each letter is in alphabetical order. For example, if $h=5$, then the partition $\{\{1,3,5\},\{2,4\}\}$ is represented by the word $ababa$.
 
 {\bf Correspondence:} For each $i$ in the $k-$th set in the word set we make correspond the $i^{th}$ character in the string to $k$.
 
 If the $i^{th}$ character in the string is $k$, put $i$ into the $k^{th}$ set in the word.
 \subsection{Combinatorics of partitions sets}
 For each partition $\lambda=(\lambda_{1},\ldots, \lambda_{k})$ we consider its Young tableux.  The tableux of $\lambda$ is an array of boxes, left justified, with $\lambda_{i}$ boxes in the $i^{th}$ row, with rows arranged from top to bottom. For example,
%If we consider the partition $\alpha=(5,3,3,1)$, its Young diagram is:
$$
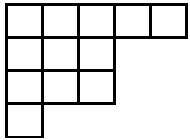
\begin{Young}
 &  &   &   & \cr
 &  & \cr
 &  &  \cr
 \cr
\end{Young}
$$
is the Young tableux of the partition $\lambda=(5,3,3,1)$ with $l(\lambda)=4$ and 
$|\lambda|=12$.  If we represent the partitions $\lambda, \mu, \gamma$ by the corresponding Young tableux, the corresponding Littlewood-Richardson coefficient $c^{\eta}_{\lambda \mu}$ represents the number of ways to fill the boxes $\eta\backslash \lambda$ with an integer $i$ in each box, so that the following conditions are satisfied:
 \begin{enumerate}
 \item The entries in any row are weakly increasing from left to right.
 \item The integer $i$ occurs exactly $\mu_{i}$ times.
 \end{enumerate} %Referencia Fulton
 
 We order the boxes by first listing the boxes in the top row, from right to left, then the boxes in the second row from right to left, and so on down the array, such that, the partition $\mu=\gamma\backslash \lambda$ consists of $\sum\, \mu_{i}$ boxes, $1 \leq i \leq n$. Young tableux are used for example to process data bases. For our purposes, we use them to pack information defined as sequences of letters, or words in an alphabet on two letters. The columns of the table would pack information in the form of sequences of bits for example, and the  rows of the table would specify features, objects or more generally meaning associated to the different words.
 
   \subsection{Correspondence between words and graphs}\label{subsection3.2}
 For each word consider its generating set $S$. Nodes and vertices of the graph are represented by elements in $S$. Now vertices in a block are in the same equivalence class (they correspond to records that are in some sense similar), and which share an edge. A complete graph is a graph in which each pair of distinct vertices is joined by an edge.
 For a graph $G$ and nonempty subset $S\subseteq V(G)$, the vertex induced subgraph denoted as $\langle X\rangle$, is the subgraph of $G$ with vertex set $S$ and edges incident to members of $S$. A colletion of subgraphs of $G$ is called a covering of the graph $G$ is every edge of $G$ is contained in one of the (not necessarily spanned) subgraphs in the collection. A subset $X$ of the vertex set is called independent set, if there is no edge between vertices.

We pass from a directed graph $D$ to a set partition by storing an array $V$ indexed by vertices, that is, $V[i]$ points to a list of neighbors of $i$. For an undirected graph $G$, $V[i]$ points to a list of heads of outgoing arcs of $i$. %In the above example, this would be ($\{\{2,3\}, \{3\}, \{4\}, \{2\} \}$).
 A multigraph $X$ is a graph $(V(X),E(X))$ with loops (edges whose endpoints are equal) and multiple edges.
 \begin{itemize}
 \item The adjacency matrix $A$ of a multigraph is a $n\times n$ matrix (where $n=|V|$) with rows and columns indexed by the elements of the vertex set and the $(x,y)-th$ entry is the number of edges connecting $x$ and $y$.
\item If the graph is directed, the matrix $A$ is symmetric and therefore all its eigenvalues are real.
\item The degree of a vertex ${\rm{deg}}(v)$ is the number of edges incident with $v$, where we count a loop with multiplicity 2. The handshaking lemma of graph theory states:
$$\sum_{v\in V}\, {\rm{deg}}(v)=2|E(x)|.$$
\item We will say that an edge has length 1, unless it is a loop, in which case we adopt the convention that it has length 2.
\item For a multigraph, a walk of length $r$ from $x$ to $y$ is a sequence $x=v_{0}, v_{1}\ldots, v_{r}=y$ with $v_{i}\in V$ and $e_{i}=(v_{i},v_{i+1})\in E$ for $i=0,1,\ldots, r-1$ and 
$$\sum_{i}l(e_{i})=r.$$
\item A path is a walk with no repeated vertex.
\item A graph is said to be connected if for any $x, y\in V$, there is a path from $x$ to $y$.
\item A graph is called  k-regular if every vertex has degree $k$. In particular all the eigenvalues of the adjacency matrix satisfy $|\lambda|\leq k$. By a Theorem due to Murty \cite{RM}, if $X$ is a $k-$regular graph, then $\lambda=k$ is an eigenvalue with multiplicity equal to the number of connected components.
\item A bipartite graph is a graph whose vertices can be partioned into two sets, in such a way that each edge has a vertex on one side and another vertex on the other side. In particular $-k$ is eigenvalue of $X$.
\item We define a $(v,b,r,k)-$combinatorial configuration as a connected bipartite graph with $v$ vertices on one side, each of them of degree $r$ and $b$ vertices on the other side, each of them of degree $k$. When $v$ and $b$ are not known, we use the notation $(r,k)-$configuration.
\item Geometrically a $(v,b,r,k)-$combinatorial configuration is an incidence structure in which there are $r$ lines through every point, $k$ points on every line,  such that through any pair of points there is at most one line.
\item For $r,k\in \mathbb{N}, r, k \geq 2$ the set
$$S_{r,k}:=\Bigg\{d\in \mathbb{N}:\, \exists\ combinatorial\ \ (v,b,r,k)-configuration$$ $$ and \ v=d\frac{k}{gcd(r,k)},\, b=d\frac{r}{gcd(r,k)}\Bigg\}$$ is a numerical semigroup,  \cite{SB}.

\end{itemize}

\subsection{Random patterned matrices: applications}\label{subsection3.3}
Graphs are graphic representations of networks, and then by the previous correspondence in section \ref{section3.1}, a network describes a language, where the symbols of the alphabet constitute the vertex set and the transformations describe the rules. So to understand which graph visualizations represent the same network, we must study topological properties of the graph such as:
\begin{itemize}
 \item Connectivity, number of connected components.
 \item Betweenness is a commonly used measure of centrality, i.e., of topological importance, both for nodes and links. It is a weighted sum of the number of shortest paths passing through a given node or a link.
 \item Network metrics such as average distance and diameter which describe the separation of nodes, which are important for evaluating the performance of routing algorithms.
 \item The largest eigenvalue $\lambda$ of the adjacency matrix describes the spectrum character of the graph topology.

 \end{itemize}
 By the correspondence from graphs to set partitions,  we can think of the matrices as words in an alphabet $A=\{0,1\}$ on two letters. Then given a tableux $T$ storing data from the entries of a matrix in $i$ rows and $j$ columns, a typical pattern problem is given as input the number of matrices $n\times n$ with a given number $2k$ of elements different from 0 and other restrictions.

 \subsection{Probability distribution of a word}\label{subsection3.4}
 %We call probability distribution of a word to the distribution followed
 Let $X_{n}$ be the number of occurrences of any given pattern in a word of length $n$ generated by a grammar or by a simple model with an alphabet of 2 letters. % can follow any limit law which we call distribution, in the sense that 
 %There exist some 
 The process of identifying the subgraph defined by a certain pattern is called re-identification. For certain patterns and some grammars one can study the characteristics of the limit curve (for large $n$) of $(k, \mathbf{P}robability(X_{n}=k))$, that is, the distribution followed by a parameter in a grammar.  
 \begin{definition}
 A re-identification method is a function that given a collection of entries in $y \in \mathcal{P}(Y)$ and some additional information from a space of auxiliary informations $A$, returns the probability that $y$ corresponds  to entries from the record with index $i \in I$.
 \end{definition}
 
 Given a Tableux $T$, we define the partition $\mathcal{P}(T)$ of $T$ to be the set of subsets of the underlying set of entries of $T$, that is, the set $\bigcup_{i,j=1}^{n,m}\,T[i,j]$ where $i$ is indexed by partitions $\lambda_{i}=(\lambda_{1},\ldots, \lambda_{k})$. This follows from the fact that there are $\lambda_{i}$ boxes in the $i^{th}$ row and $j$ is indexed by partitions $\gamma_{k}=\#\,\{\lambda_{i}=k\}$ corresponding to the number of times the multiplicity corresponding to the integer $k$ is realized.

We say that the entries $s\in \mathcal{P}(Y)$ which represent occurrences in a probability space are linked to a collection of indices $J\subset I$ if the probabilities that are returned by the reidentification method take non-zero values over the indices $J$ and are zero on the complement $I/J$. In a regular situation, a possible non zero value for the re-identification method over $J$ is then $1/|J|$. 
 
Consider the objective probability distribution corresponding to the Horn's problem studied in \cite{BM} in the context of algebraic codes. %following reidentification method in the sense of Stokes-Torra in \cite{ST}: %probabilistic model
 
 \begin{eqnarray}
r:\, \mathcal{P}(T)\times A \rightarrow [0,1]^{n} \\
 (y,a) \rightarrow \mathbf{P}(y\, corresponds\, to\, entries\, from\, T[i,j]:\, i\in I, j\in J),
 \end{eqnarray}
 that, to a pair $(y,a)\in \mathcal{P}(T)\times A$, where $A$ is an information space, associates the probability that the integer $i$ occurs exactly $\lambda_{i}$ times, and $j$ occurs exactly $\gamma_{j}$ times, that is:
 $$\mathbf{P}(\bigcup_{i,j=1}^{n,m}T[i,j]=k)=\mathbf{P}(\{\mu_{k}=k\}).$$
 
 Let $J\subseteq I$, then we denote the projection of the table on the set of indeces $j$ by $T[J]$. Two sets of indexes are related, if the probabilities returned by the reidentification method are non-zero and are 0 on the complement $I\backslash J$.

 Horn gives an inductive procedure to produce set of triples $(I,J,K)\subset \{0,1,\ldots, n\}$. A set of indexes is associated to a partition in the following way $\lambda=(i_{r}-r,\ldots, i_{1}-1)$, see appendix of \cite{BM}. 
 Then the algorithm produces triples of partitions $(\lambda,  \gamma, \mu)$ that are admissible for the Horn problem, that is, they are in correspondence with eigenvalues of Hermitian matrices $A, B, C$ such that the sum of any two of them is equal to the third.

 A probabilistic record linkage is a mathematical model based on a probabilistic model that computes the probability of a particular coincidence $\gamma$ condicionated by the existence of a match. Namely, the ordinary bivariate generating function $$r(y,a)[i]=\mathbf{P}(Match|\, \gamma(y,x_{i})),$$ computes the probability of a particular coincidence pattern $\gamma$ condicionated by the existence of a match. This can be translated in terms of colors and indices.

 \begin{theorem} Given a partition $\gamma$, the probability $$\mathbf{P}(\gamma_{k}=k, 1\leq k \leq n)=\frac{c^{\gamma}_{\lambda, \mu}}{3\cdot 2^{n+2}},$$ where $c^{\gamma}_{\lambda, \mu}$, is the Littlewood-Richardson coefficient associated to the partitions $\gamma, \lambda, \mu$.

 \end{theorem}
 {\it Proof.}  Given a partition $\gamma$, the probability $\mathbf{P}(\gamma_{k}=k, 1\leq k \leq n)$,%=\frac{c^{\gamma}_{\lambda, \mu}}{n^{k+1}}$, 
 means the probability that the integer $k$ occurs exactly $\gamma_{k}$ times in the box indexed by partitions $\lambda, \mu$, where $(\gamma, \lambda, \mu)$ are the partitions admissible for the Horn problem. The Littlewood-Richardson coefficient $c^{\gamma}_{\lambda, \mu}$ represents the number of ways to fill the boxes $\mu\backslash \lambda$ with integer $i$ in each box, so that the following conditions are satisfied:
 \begin{enumerate}
 \item The entries in any row are weakly increasing from left to right.% \\
 \item The integer i occurs exactly $\gamma_{i}$ times.
 \end{enumerate}

We order the boxes by first listing them in the top row, from right to left, then the boxes in the second row from right to left, and so on down the array.  Since the number of partitions which fit a $d\times (n-d)$ rectangle are in bijection with 0-1 strings of $(n-d)\, 0's$ and $d\, 1's$, we must divide by $2^{n}$. Now, since the Littlewood-Richardson coefficients are invariant under the action of the dihedral group $\mathbb{Z}_{2}\times S_{3}$, \cite{PV}, we must further divide by the order of this group, that is 12. 

\cqd

\begin{remark} 
{\rm {Combinatorial formulae for the Littlewood-Richardson coefficients have been given in \cite{KT}.
To each triplet $(\alpha, \gamma, \beta)$ of partitions we can associate a polytope in which the number of lattice points is the corresponding Littlewood-Richardson coefficient, (Tao and Knutsen use the honeycomb model}}.
\end{remark}

\begin{remark} {\rm {Observe that %$\mathbb{Z}^{3}$ is a probability space for 
the network described by triples of partitions $(\lambda,\mu, \gamma)$ for which the corresponding Littlewood-Richardson coefficient $c^{\lambda}_{\mu, \gamma}>0$ is very much connected with the polytope of matrices admissible for Horn's problem. The network describes the language where words are points in the affine manifold associated to the polytope}}.
\end{remark}

Horn gives an inductive procedure to produce a set of triples $(I,J,K)\subset \{0,1,\ldots, n\}$, see \cite{Fu}. The partition $\lambda=(i_{r}-r,\ldots, i_{1}-1)$ is associated with a set of indices.% see appendix of \cite{BM}. 
 Then the algorithm produces triples of partitions $(\lambda,  \gamma, \mu)$ that are admissible for the Horn problem, that is, they are in correspondence with eigenvalues of Hermitian matrices $A, B, C$ such that the sum of any two of them is the third.
 
 $$U^{n}_{r}=\{(I,J,K)|\, \sum_{i\in I} i+ \sum_{j\in J} j=\sum_{k\in K}k+r(r+1)/2\},$$

$$T^{n}_{r}=\{(I,J,K)\in U^{n}_{r}|\, for\ all\ p<r \ and \ all \ (F,G,H) \in T^{r}_{p},$$ $$ \sum_{f\in F}i_{f}+\sum_{g\in G}j_{g}\leq \sum_{h\in H}k_{h}+p(p+1)/2\}.$$

We have implemented this algorithm using Python: this involves calculate and iterate through $r$-combination of $n$ element.
The running time is $O({n \choose r}^3)$.
The next table shows the values of the indices for parameters~$1\leq n, r \leq 4$.
\begin{tiny}

\begin{center}
\begin{tabular}{ | l | p{4cm} | p{4cm} | }

\hline
$(n, r)$   & $U^{n}_{r}$ &  $T^{n}_{r}$ \\
\hline
\hline
(2, 1) & $(\{1\}, \{1\}, \{1\})$, $(\{1\}, \{2\}, \{2\})$, $(\{2\}, \{1\}, \{2\})$ & $(\{1\}, \{1\}, \{1\})$, $(\{1\}, \{2\}, \{2\})$, $(\{2\}, \{1\}, \{2\})$ \\
\hline
(3, 1) & $(\{1\}, \{1\}, \{1\})$, $(\{1\}, \{2\}, \{2\})$, $(\{1\}, \{3\}, \{3\})$, $(\{2\}, \{1\}, \{2\})$, $(\{2\}, \{2\}, \{3\})$, $(\{3\}, \{1\}, \{3\})$ & $(\{1\}, \{1\}, \{1\})$, $(\{1\}, \{2\}, \{2\})$, $(\{1\}, \{3\}, \{3\})$, $(\{2\}, \{1\}, \{2\})$, $(\{2\}, \{2\}, \{3\})$, $(\{3\}, \{1\}, \{3\})$ \\
\hline
(3, 2) & $(\{1, 2\}, \{1, 2\}, \{1, 2\})$, $(\{1, 2\}, \{1, 3\}, \{1, 3\})$, $(\{1, 2\}, \{2, 3\}, \{2, 3\})$, $(\{1, 3\}, \{1, 2\}, \{1, 3\})$, $(\{1, 3\}, \{1, 3\}, \{2, 3\})$, $(\{2, 3\}, \{1, 2\}, \{2, 3\})$ & $(\{1, 2\}, \{1, 2\}, \{1, 2\})$, $(\{1, 2\}, \{1, 3\}, \{1, 3\})$, $(\{1, 2\}, \{2, 3\}, \{2, 3\})$, $(\{1, 3\}, \{1, 2\}, \{1, 3\})$, $(\{1, 3\}, \{1, 3\}, \{2, 3\})$, $(\{2, 3\}, \{1, 2\}, \{2, 3\})$ \\
\hline
\hline
(4, 1)  & $(\{1\}, \{1\}, \{1\})$, $(\{1\}, \{2\}, \{2\})$, $(\{1\}, \{3\}, \{3\})$, $(\{1\}, \{4\}, \{4\})$, $(\{2\}, \{1\}, \{2\})$, $(\{2\}, \{2\}, \{3\})$, $(\{2\}, \{3\}, \{4\})$, $(\{3\}, \{1\}, \{3\})$, $(\{3\}, \{2\}, \{4\})$, $(\{4\}, \{1\}, \{4\})$ & $(\{1\}, \{1\}, \{1\})$, $(\{1\}, \{2\}, \{2\})$, $(\{1\}, \{3\}, \{3\})$, $(\{1\}, \{4\}, \{4\})$, $(\{2\}, \{1\}, \{2\})$, $(\{2\}, \{2\}, \{3\})$, $(\{2\}, \{3\}, \{4\})$, $(\{3\}, \{1\}, \{3\})$, $(\{3\}, \{2\}, \{4\})$, $(\{4\}, \{1\}, \{4\})$ \\
\hline
(4, 2) & $(\{1, 2\}, \{1, 2\}, \{1, 2\})$, $(\{1, 2\}, \{1, 3\}, \{1, 3\})$, $(\{1, 2\}, \{1, 4\}, \{1, 4\})$, $(\{1, 2\}, \{1, 4\}, \{2, 3\})$, $(\{1, 2\}, \{2, 3\}, \{1, 4\})$, $(\{1, 2\}, \{2, 3\}, \{2, 3\})$, $(\{1, 2\}, \{2, 4\}, \{2, 4\})$, $(\{1, 2\}, \{3, 4\}, \{3, 4\})$, $(\{1, 3\}, \{1, 2\}, \{1, 3\})$, $(\{1, 3\}, \{1, 3\}, \{1, 4\})$, $(\{1, 3\}, \{1, 3\}, \{2, 3\})$, $(\{1, 3\}, \{1, 4\}, \{2, 4\})$, $(\{1, 3\}, \{2, 3\}, \{2, 4\})$, $(\{1, 3\}, \{2, 4\}, \{3, 4\})$, $(\{1, 4\}, \{1, 2\}, \{1, 4\})$, $(\{1, 4\}, \{1, 2\}, \{2, 3\})$, $(\{1, 4\}, \{1, 3\}, \{2, 4\})$, $(\{1, 4\}, \{1, 4\}, \{3, 4\})$, $(\{1, 4\}, \{2, 3\}, \{3, 4\})$, $(\{2, 3\}, \{1, 2\}, \{1, 4\})$, $(\{2, 3\}, \{1, 2\}, \{2, 3\})$, $(\{2, 3\}, \{1, 3\}, \{2, 4\})$, $(\{2, 3\}, \{1, 4\}, \{3, 4\})$, $(\{2, 3\}, \{2, 3\}, \{3, 4\})$, $(\{2, 4\}, \{1, 2\}, \{2, 4\})$, $(\{2, 4\}, \{1, 3\}, \{3, 4\})$, $(\{3, 4\}, \{1, 2\}, \{3, 4\})$ & $(\{1, 2\}, \{1, 2\}, \{1, 2\})$, $(\{1, 2\}, \{1, 3\}, \{1, 3\})$, $(\{1, 2\}, \{1, 4\}, \{1, 4\})$, $(\{1, 2\}, \{2, 3\}, \{2, 3\})$, $(\{1, 2\}, \{2, 4\}, \{2, 4\})$, $(\{1, 2\}, \{3, 4\}, \{3, 4\})$, $(\{1, 3\}, \{1, 2\}, \{1, 3\})$, $(\{1, 3\}, \{1, 3\}, \{1, 4\})$, $(\{1, 3\}, \{1, 3\}, \{2, 3\})$, $(\{1, 3\}, \{1, 4\}, \{2, 4\})$, $(\{1, 3\}, \{2, 3\}, \{2, 4\})$, $(\{1, 3\}, \{2, 4\}, \{3, 4\})$, $(\{1, 4\}, \{1, 2\}, \{1, 4\})$, $(\{1, 4\}, \{1, 3\}, \{2, 4\})$, $(\{1, 4\}, \{1, 4\}, \{3, 4\})$, $(\{2, 3\}, \{1, 2\}, \{2, 3\})$, $(\{2, 3\}, \{1, 3\}, \{2, 4\})$, $(\{2, 3\}, \{2, 3\}, \{3, 4\})$, $(\{2, 4\}, \{1, 2\}, \{2, 4\})$, $(\{2, 4\}, \{1, 3\}, \{3, 4\})$, $(\{3, 4\}, \{1, 2\}, \{3, 4\})$ \\
\hline
(4, 3) & $(\{1, 2, 3\}, \{1, 2, 3\}, \{1, 2, 3\})$, $(\{1, 2, 3\}, \{1, 2, 4\}, \{1, 2, 4\})$, $(\{1, 2, 3\}, \{1, 3, 4\}, \{1, 3, 4\})$, $(\{1, 2, 3\}, \{2, 3, 4\}, \{2, 3, 4\})$, $(\{1, 2, 4\}, \{1, 2, 3\}, \{1, 2, 4\})$, $(\{1, 2, 4\}, \{1, 2, 4\}, \{1, 3, 4\})$, $(\{1, 2, 4\}, \{1, 3, 4\}, \{2, 3, 4\})$, $(\{1, 3, 4\}, \{1, 2, 3\}, \{1, 3, 4\})$, $(\{1, 3, 4\}, \{1, 2, 4\}, \{2, 3, 4\})$, $(\{2, 3, 4\}, \{1, 2, 3\}, \{2, 3, 4\})$  & $(\{1, 2, 3\}, \{1, 2, 3\}, \{1, 2, 3\})$, $(\{1, 2, 3\}, \{1, 2, 4\}, \{1, 2, 4\})$, $(\{1, 2, 3\}, \{1, 3, 4\}, \{1, 3, 4\})$, $(\{1, 2, 3\}, \{2, 3, 4\}, \{2, 3, 4\})$, $(\{1, 2, 4\}, \{1, 2, 3\}, \{1, 2, 4\})$, $(\{1, 2, 4\}, \{1, 2, 4\}, \{1, 3, 4\})$, $(\{1, 2, 4\}, \{1, 3, 4\}, \{2, 3, 4\})$, $(\{1, 3, 4\}, \{1, 2, 3\}, \{1, 3, 4\})$, $(\{1, 3, 4\}, \{1, 2, 4\}, \{2, 3, 4\})$, $(\{2, 3, 4\}, \{1, 2, 3\}, \{2, 3, 4\})$ \\
\hline

\end{tabular}
\end{center}

\end{tiny}

 %So partition $\mu\backslash \lambda$ consists of $\sum_{i=1}^{n} \gamma_{i}$ boxes.
\subsubsection{Random patterned matrices}
A perfect secret sharing scheme $S$ for a finite graph $G$ is a collection of random variables $\xi_{v}$ for each $v\in V$ and $\xi_{s}$ (the secret) with a joint distribution so that:
\begin{enumerate}
\item Two random variables $\xi_{v}$ and $\xi_{w}$ together recover the value of $\xi_{s}$ if $vw$ is and edge in G;
\item For any independent set $A$, the $\xi_{s}$ and the collection of variables $\{\xi_{v}:\, v\in A\}$ are statiscally independent.
\end{enumerate}

A sequence or bi-sequence of variables $\{x_{i}: i\geq 0\}$ or $\{x_{ij}: i, j \geq 1\}$ will be called an input sequence. Let $\mathbb{Z}$ be the set of all integers and let $\mathbb{Z}_{+}$ denote the set of all non-negative integers. Let
$$L_{n}:\, \{1,2,\ldots, n \}^{2}\rightarrow \mathbb{Z}^{d}, \, n\geq 1, \ d=1, 2$$ be a sequence of functions such that $L_{n+1}(i,j)=L_{n}(i,j)$ whenever
$1\leq i, j \leq n$.
%words--matrices
%words--partitions
 A $k-$regular partition of $n~(k>1)$ is a non-increasing sequence of positive integers whose sum is $n$, with the condition that no summand is divisible by $k$. We shall write $L_{n}=L$ and call it the {\it link} function and we write $\mathbb{Z}^{2}_{+}$ as the common domain of $\{L_{n}\}$. Patterned matrices are those defined by $X_{n}=((x_{L(i,j)}))$.
 
 Any function $\pi: \{0,1,2,\ldots,h\}\rightarrow \{1,2,\ldots, n\}$ with $\pi(0)=\pi(h)$ is called a circuit of {\it length} $h$. We say that two circuits $\pi_{1}$ and $\pi_{2}$ are equivalent if and only if their $L$ values respectively match at the same locations, that is,
{\small{$$L(\pi_{1}(i-1),\pi_{1}(i))=L(\pi_{1}(j-1),\pi_{1}(j))\iff L(\pi_{2}(i-1),\pi_{2}(i))=L(\pi_{2}(j-1),\pi_{2}(j)).$$}} 
 
 %\begin{defi}
 \begin{enumerate}
 \item A circuit is matched if all $L-$values $L(\pi(j-1), \pi(j))$ are repeated more than once.
 \item If $L-$values are repeated exactly twice, then it is called pair matched.
 \item If the $L-$values are repeated with the same color, then it is color matched.
 \end{enumerate}
If we work with an alphabet of two letters, we color them with colors $a$ and $b$ respectively.

 A word is said to be catalan if it is pair-matched and deleting all double letters leads to the empty word. For example $abba$ is catalan and $abab$ is not, \cite{Ba}.
 
 Let $w$ be a catalan word of length $2k$. Let $S$ denote the set of all generating vertices of $w$. Then for all $j\notin S$, there exists a unique $i \in S$ such that $i<j$ and $\pi(j)=\pi(i)$ for all $\pi\in \pi^{*}(w)$.

Let {\small{$$CW_{A}(2)=\{{\rm{all \ \, words\  of \  length\  k \  which \ are \  pair \  matched \   (within \  the  \ same \   color)}}. \}$$}}
The equivalence class corresponding to $w$ and the set of pair matched noncolored words will be denoted by

$$\Pi(w)=\{\pi:\, w[i]=w[j] \iff L(\pi(i-1),\pi(i))=L(\pi(j-1),\pi(j))\}.$$

 A partition $(\lambda_{1},\ldots, \lambda_{r})$ is $k-$regular if no part $\lambda_{i}$, $1\leq i \leq r$ is divisible by $k$. In classical representation theory, $k-$regular partitions of $n$ label irreducible $k-$modular representations of the symmetric group $S_{n}$ when $k$ is prime.
  
 Any equivalence class can be indexed by a partition of $\{1,2,\ldots, h\}$. Each block of a given partition identifies the positions where the $L-$matches take place. We can label these partitions by words of letters where the first occurrence of each letter is in alphabetical order. For example, if $h=5$ then the partition $\{{1,3,5}, {2,4}\}$ is represented by the word $ababa$.

 {\bf Example:} For a catalan word $w$ of lenght $2k$ we define:
{$$\pi_{1}^{*}(w)=\{\pi: w[i]=w[j]|\, \pi(i-1)+\pi(i)=\pi(j-1)+\pi(j), \pi(i-1)+\pi(i)\leq n+1\}.$$}
 For any $j$ not necessarily in $S$, let us denote by $\phi(j)$ the unique vertex such that $\phi(j)\in S, \, \phi(j)\leq j$ and $\pi(j)=\pi(\phi(j))\ \ \forall \pi \in \pi^{*}(w)$.

It follows that $\frac{\sharp \pi^{*}_{1}(w)}{n^{1+k}}$ is the Riemann sum:
$$I_{w}(v_{s})=I(v_{\phi(i-1)}+v_{\phi(i)})\leq 1, i\in S-\{0\}\, \, over \, \, [0,1]^{k+1}.$$

Let us define $v_{i}=\frac{\pi(i)}{n},$ $U_{n}=\{\frac{1}{n},\ldots, \frac{n-1}{n},1\}$ and $v_{s}=\{v_{i}:\, i\in S\}.$

It converges to the integral:
$$lim_{n\rightarrow \infty}\frac{1}{n^{1+k}}\sharp\, \pi_{1}^{*}(w)=\int_{[0,1]^{k+1}}I_{w}(v_{s})dv_{s}$$

Let $w$ be the word $aa$, then  evaluating the integral $I_{w}$ leads to:
$$p_{u}(w)=\int_{v_{0}+v_{1}\leq 1}dv_{0}v_{1}=\int_{0}^{1}(1-v_{0})dv_{0},$$ and hence $Q_{aa}(x)=1-x,$ where $x$ is the probability that two nodes share an edge. Thus $p_{u}(aa)=Q_{aa}(2)=\frac{1}{2}.$

Let $w$ be the word $abba$, then
$$p_{u}(w)=\int_{[0,1]^{3}}\ldots \int I(v_{0}+v_{1}\leq 1, v_{1}+v_{2}\leq 1)dv_{0}dv_{1}dv_{2}=$$
$$ \int_{0}^{1}\int_{0}^{1-v_{0}}(1-v_{1})dv_{1}dv_{0}=\int_{0}^{1}\frac{1-v_{0}^{2}}{2}dv_{0}.$$

Hence $Q_{abba}(x)=\frac{1-x^{2}}{2}$. Thus $p_{u}(aabb)=\frac{1}{3}.$

\begin{remark}
{\rm We can make a correspondence from the space of events/ information space to sequences of words/strings in such a way that it is possible to compute the probability of a particular coincidence conditionated by the existence of a match by means of the generating functions of the corresponding sets of words.
\item In particular the integral $p_{u}(w)$ is an example of a reidentification method $r(y,a)[i]$ where $i$ is an index in the set of indexes, by identifying $y=u$ and $w=a$}.
\end{remark}

\end{document}